\documentclass[conference]{IEEEtran}
\IEEEoverridecommandlockouts
\usepackage{cite}
\usepackage{amsmath,amssymb,amsfonts}
\usepackage{algorithmic}
\usepackage{graphicx}
\usepackage{textcomp}
\usepackage{xcolor}
\usepackage{booktabs}
\usepackage{multirow}
\def\BibTeX{{\rm B\kern-.05em{\sc i\kern-.025em b}\kern-.08em
    T\kern-.1667em\lower.7ex\hbox{E}\kern-.125emX}}
\newcommand{\techname}{GAA}

\begin{document}

\title{A Gradient based Adversarial Augmentation Framework for Mitigating Unlearnable Datasets} 

\author{
\IEEEauthorblockN{Long Dang, Satpal Sura, Kaiqi Xiong }
\IEEEauthorblockA{\textit{ICNS Lab and Cyber Florida} \\
\textit{University of South Florida}\\
FL, USA \\
longdang@usf.edu, satpal@usf.edu, xiongk@usf.edu}
}

\maketitle

\begin{abstract}
Unlearnable data methods (ULD) protect public data from unauthorized model training by adding imperceptible perturbations that create a learning shortcut. A deep neural network can exploit the shortcut learning path to minimize training loss without learning generalizable features. Intuitively, finding adversarial versions of perturbed images that can substantially increase the model's loss can force the model to learn generalizable features. Existing data augmentation based attacks rely on heuristic search strategies/random sampling over a curated augmentation set to identify adversarial augmentation operations, an approach that often does not scale when the number of candidate operations is large. To address the challenge of a large search space, we propose {\techname}, a gradient based adversarial augmentation framework that learns an augmentation policy via stochastic gradient ascent. {\techname} leverages a differentiable data augmentation pipeline to efficiently search a set of up to 18 augmentation techniques, from which a grid search selects a protection-specific subset. The augmentation subset is trained to maximize a surrogate model's classification loss. We introduce a perceptual-plus-pixel regularization to learn operation parameters that preserve image semantics. Under a controlled ablation, {\techname} consistently outperforms an equivalent random sampling baseline (80.0\% vs. 77.29\% clean test accuracy on EMIN-protected CIFAR-10 for ResNet-18 and 83.28\% vs. 66.45\% on NTGA-protected CIFAR-10 for MobileNet-V2), demonstrating that gradient guidance drives the clean test accuracy recoverability rather than the expanded search space. {\techname} offers an effective and scalable adversarial augmentation framework for stress-testing ULD methods, a critical tool for developing robust ULD methods.
\end{abstract}
\section{Introduction}

Deep learning models increasingly rely on large-scale datasets scraped from the internet, prompting the development of unlearnable data methods (ULD) that prevent deep neural networks (DNNs) from extracting meaningful representations without data owner consent~\cite{li2025surveyunlearnabledata}. Representative ULD methods include error-minimizing noise (EMIN)~\cite{EM}, autoregressive perturbations (AR)~\cite{AR}, targeted adversarial poisoning (TAP)~\cite{fowl2021adversarial}, neural tangent generalization attacks (NTGA)~\cite{yuan2021neural}, self-ensemble protection (SEP)~\cite{DBLP:conf/iclr/ChenYC0QWH23}, and DeepConfuse (DC)~\cite{deepconfuse}, each of which allows embedding a distinct, protection-specific shortcut into the training data. Existing studies, including nonlinear transformation~\cite{hapuarachchi2024nonlinearTransformationsAgainst}, ISS~\cite{liu2023ImageShortcutSqueezing}, and UEraser~\cite{qin2023learning}, attempt to leverage data augmentation (DA) techniques as an effective attack to eliminate the learning shortcuts introduced by the ULD defenses. Although these attack studies can restore the model's ability to learn generalizable features, they often assume a limited search space. In ISS~\cite{liu2023ImageShortcutSqueezing}, the DA space consists of simple compression techniques including Grayscale, JPEG and Bit Depth Rate (BDR), making it easy for data owners to leverage these techniques to break the ISS attack~\cite{dang2025enhancing}. Nonlinear transformation~\cite{hapuarachchi2024nonlinearTransformationsAgainst} counters ULD by sequentially searching for the best combination of multiple transformations given by a five-element search space. Finding the best sequence of an exact length of three requires testing every possible combination in a space of ($5^3$) possibilities. Thus, their proposed search algorithm is limited by the size of the search space. Although UEraser~\cite{qin2023learning} improves the generalization of DNNs against multiple ULD datasets without reducing clean test accuracy, it significantly increases the training costs from 30 to 200 epochs for Resnet-18 on CIFAR-10 dataset. Furthermore, if we consider the need to fine-tune the hyperparameters for each DA operation for optimal performance, the training process might never end~\cite{cubuk2018autoaugment}. See Table~\ref{tab:summary_data_aug} for a list of transformations used in the DA based attacks. At the same time, a line of research studies how to efficiently learn augmentation strategies for image classification tasks~\cite{yang2022survey}. To enhance the robustness of DA based attacks against adaptive ULD defenses and discover novel combinations from a broad space of DAs, in this study, we focus on searching for adversarial DA augmentations from a broad set of standard DA techniques in an automated manner.
\begin{table}[htbp]
\caption{The search space of existing data augmentation based attacks.}
    \begin{tabular}{|p{3.1cm}| p{5.0cm}|}
    \hline
      \textbf{Attacks} & \textbf{Search space} \\
    \hline
    Nonlinear transformations.~\cite{hapuarachchi2024nonlinearTransformationsAgainst}  & Threshold binary, threshold binary inverse, color channel manipulation, erode, and Gaussian blur \\
     \hline
     ISS~\cite{liu2023ImageShortcutSqueezing} & Grayscale, JPEG and Bit Depth Rate (BDR) \\
    \hline
    UEraser~\cite{qin2023learning} & PlasmaTransform, TrivialAugment, and ChannelShuffle \\
     \hline
     Ours ({\techname}) & We expand the DA search space beyond previous studies~\cite{hapuarachchi2024nonlinearTransformationsAgainst, liu2023ImageShortcutSqueezing, qin2023learning} by utilizing the DA set of 16 standard image transformation techniques listed in AutoAugment~\cite{cubuk2018autoaugment}. We further add Gaussian blur and Grayscale to ensure generating diverse augmented images. In summary, the augmentation search space has up to 18 operations. A protection-specific operative subset is selected via grid search (Sec.~\ref{sec:search_space}){$\dagger$}. \\
     \hline
    \end{tabular}
    \label{tab:summary_data_aug}
    \raggedright
    \footnotesize
    $^{\dagger}$Selected operative subset: base+blur (17 operations) for EMIN, AR, TAP, SEP, and DC; base+blur+grayscale (18 operations) for NTGA.
\end{table}
Mitigating unlearnable examples with data augmentation poses several challenges to both ensure efficiency and recover clean test accuracy. Existing data augmentation based defenses such as UEraser~\cite{qin2023learning} restrict their search to a small, curated pool of handpicked transformations, limiting the diversity of augmentations available to disrupt learning shortcuts. Automated augmentation policy search frameworks such as Faster AutoAugment~\cite{hataya2020faster}, while effective for improving generalization, have instead been optimized to match the distribution of clean data rather than to adversarially maximize classification loss, making them poorly suited to mitigating unlearnable perturbations without adaptation. Adversarial training~\cite{madry2018towards}, though partially effective, remains confined to the same $\ell_p$-bounded threat model assumed by the attacker, so matching the attacker's perturbation strength comes at a proportional cost to clean accuracy and requires solving a min-max optimization. Overcoming these challenges requires a search process that encompasses a wide range of augmentation operations, adaptively identifies the most effective transformations, and operates efficiently without the need for manual tuning or frequent human intervention required by prior defense. Because unlearnable perturbations function as a shortcut that lets a model minimize training loss without learning generalizable features, a transformation that substantially increases a model's loss on the perturbed data is one that likely disrupts this shortcut, forcing the model back toward learning generalizable features.
In this paper, we propose a gradient based adversarial augmentation framework ({\techname}). Our framework learns an adversarial augmentation policy that transforms protected images to maximize a surrogate model's classification loss. The loss function includes perceptual and pixel-space regularization terms to preserve the transformed image’s semantic content. During the policy training, at each stage of a sub-policy, candidate operations are combined via a softmax-weighted sum, making the policy differentiable end-to-end. The policy differentiability enables its parameters to be learned via stochastic gradient ascent~\cite{hataya2020faster}. While Faster AutoAugment~\cite{hataya2020faster} minimizes the distributional distance between augmented and clean images to improve generalization, {\techname} maximizes classification loss to suppress unlearnable protections. {\techname} is similarly related to, but distinct from, UEraser~\cite{qin2023learning}. UEraser solves an inner maximization over a small, curated pool of three augmentations via random sampling to find augmentation policies that are capable of distorting protected perturbations. {\techname} solves the same maximization directly via gradient ascent over a 18 operations search space, and utilizes UEraser's two-phase training schedule only for applying the resulting policy.

We evaluate {\techname} against a random sampling ablation (RS) and the fixed-policy UEraser across six protected CIFAR-10 datasets (EMIN~\cite{EM}, AR~\cite{AR}, NTGA~\cite{yuan2021neural}, SEP~\cite{DBLP:conf/iclr/ChenYC0QWH23}, TAP~\cite{fowl2021adversarial}, and DC~\cite{deepconfuse}) and three target architectures (ResNet-18~\cite{he2016deep}, ResNet-50~\cite{he2016deep}, MobileNetV2~\cite{sandler2018mobilenetv2}). Under a matched comparison where the policy is evaluated on the same EMIN-protected data and a subset of the EMIN dataset is used to train its surrogate, {\techname} consistently outperforms RS on EMIN dataset in terms of clean test accuracy (80.0\% vs. 77.29\%) and NTGA dataset (85.88\% vs. 80.28\%) when the target model is ResNet-18, demonstrating that gradient based search finds more effective adversarial augmentations than RS when the policy space, surrogate, and training pipeline are held fixed. This advantage does not persist under cross-data protection transfer scenario, where the policy is applied to protection methods unseen during search. Specifically, RS outperforms {\techname} on the AR and DC, indicating that {\techname}'s learned policy specializes to the shortcut structure of its training-time surrogate rather than generalizing broadly across protection methods. In Sec.~\ref{sec:eval_cross_data_protection_transfer}, we discuss this specialization in the challenging cross-data protection case.

Our main contribution is summarized as follows:
\begin{itemize}
\item We propose a gradient based adversarial augmentation framework ({\techname}) over a set of up to 18 DA techniques for the challenge of unlearnable datasets. The gradient based framework can autonomously discover multiple combinations of augmentation techniques that maximize a surrogate model's loss function. 
\item We introduce a novel surrogate model's loss function over the space of the DA techniques for mitigating unlearnable data protections. 
The regularized loss function finds augmentation combinations that both increase the value of the classification loss function defined by the labels (i.e., cross-entropy) while constraining the semantic drift via the pixel-space and perceptual penalties. We leverage a gradient estimation adapted from Faster AutoAugment~\cite{hataya2020faster} to compute the gradients of the regularized loss function w.r.t non-differentiable augmentation techniques.
\item We demonstrate the effectiveness, cross data protection, and cross model transferability of the learned policy on CIFAR-10 across six ULD methods and three DNN architectures. We evaluate {\techname}'s robustness by discussing how {\techname} can defend an adaptive data protection approach.
\end{itemize}
\section{Related work}
\subsection{Unlearnable examples}
ULD methods prevent DNNs from effectively learning from unauthorized public datasets by adding imperceptible perturbations to the training images. EMIN~\cite{EM} generates imperceptible perturbations using a min-min optimization framework, which alternatively optimizes the perturbations and a surrogate DNN by minimizing the cross-entropy loss function. Since EMIN is ineffective against adversarial training, Robust Error-Minimizing~\cite{2022_FuRobust} uses a min-min-max formulation to generate stronger unlearnable examples. CUDA~\cite{sadasivan2023cuda} embeds class-specific perturbations via frequency-domain convolution kernels rather than costly iterative optimization. One-Pixel-Shortcut~\cite{wu2023onepixel} shows that perturbing even a single pixel of each training image can degrade a model's accuracy on clean datasets. AR~\cite{AR} proposes a generic perturbation that can be generated without access to the broader dataset. The perturbation generated by AR can be applied across different datasets and used to poison different architectures.
\subsection{Automated data augmentations}
While many automated DA techniques~\cite{cubuk2018autoaugment,lim2019fast} is used for improving model generalization on unseen data, applying them specifically to counter unlearnable examples remains comparatively unexplored. Automated Data Augmentation (AutoDA)~\cite{cubuk2018autoaugment} requires substantial computational time, while Fast AutoAugment~\cite{lim2019fast} uses an efficient density matching algorithm to model the distribution of augmented and clean training samples. AutoDA methods further improve scalability by approximating the gradient of augmentation hyperparameters, making the search differentiable and jointly optimizable with the model parameters to reduce the training loss. However, EMIN perturbation noise is crafted by alternatively optimizing the noise and a surrogate model to minimize the training loss. Thus, applying an AutoDA policy over EMIN unlearnable examples is likely to reduce the training loss, enhancing unlearnability. UEraser~\cite{qin2023learning} recovers the generalization capability of DNNs by proposing a training algorithm with an adversarial oversampling technique. In each training iteration, it samples $K$ candidates from a set of three DAs and only evaluates the model's classification capability using the augmented image that maximizes the training loss. Unlike UEraser~\cite{qin2023learning}, we decouple the DA search process from the training algorithm.  We introduce a novel adversarial objective function to learn multiple combinations of DAs for the challenge of ULD methods. We directly optimize the objective function by leveraging the differentiable DA architecture and its gradient estimation of non-differentiable DA techniques~\cite{hataya2020faster}. We utilize UEraser's oversampling technique to improve the effectiveness of the DA policy found by ~\techname. 
\section{Methods}\label{sec:method}
\subsection{Gradient based Adversarial Augmentation (\techname)}
DA based attacks on ULD datasets proceed in two conceptual steps: identifying which training examples carry a shortcut, and disrupting that shortcut with a series of augmentation operations having sufficient distortion capability. Existing methods differ in how they perform identification. ISS~\cite{liu2023ImageShortcutSqueezing} and UEraser~\cite{qin2023learning} detect unlearnable examples by observing anomalously low training loss. While ISS relies on an image analysis of the perturbation to select appropriate compression techniques, UEraser utilizes a curated set of augmentations with high distortion capability that yields high training loss. Unlike ISS and UEraser, nonlinear transformation~\cite{hapuarachchi2024nonlinearTransformationsAgainst} applies a fixed sequence of transformations without an explicit identification step. {\techname} identifies protected samples using the same low-training-loss pattern as UEraser and ISS, but automates the identification process via fitting a surrogate model on the protected dataset. The fitted surrogate model can predict protected examples precisely, driving its training loss toward zero.

Fitting a surrogate model on the protected dataset causes it to depend on the shortcut for its predictions. Consequently, any transformation that disrupts the shortcut's pixel pattern will substantially increase the surrogate model's loss, while transformations that leave the pattern intact will not. Searching for the transformation that maximizes the surrogate's loss therefore serves as a proxy for searching for the transformation that most effectively destroys the shortcut. Applying this transformation over unlearnable examples removes the shortcut path that substantially decreases the training loss, forcing the model to learn generalizable features and allowing clean test accuracy to recover.
\subsubsection{Architecture}\label{sec:architecture}
\begin{figure*}[t]
    \centering
    \includegraphics[width=\textwidth]{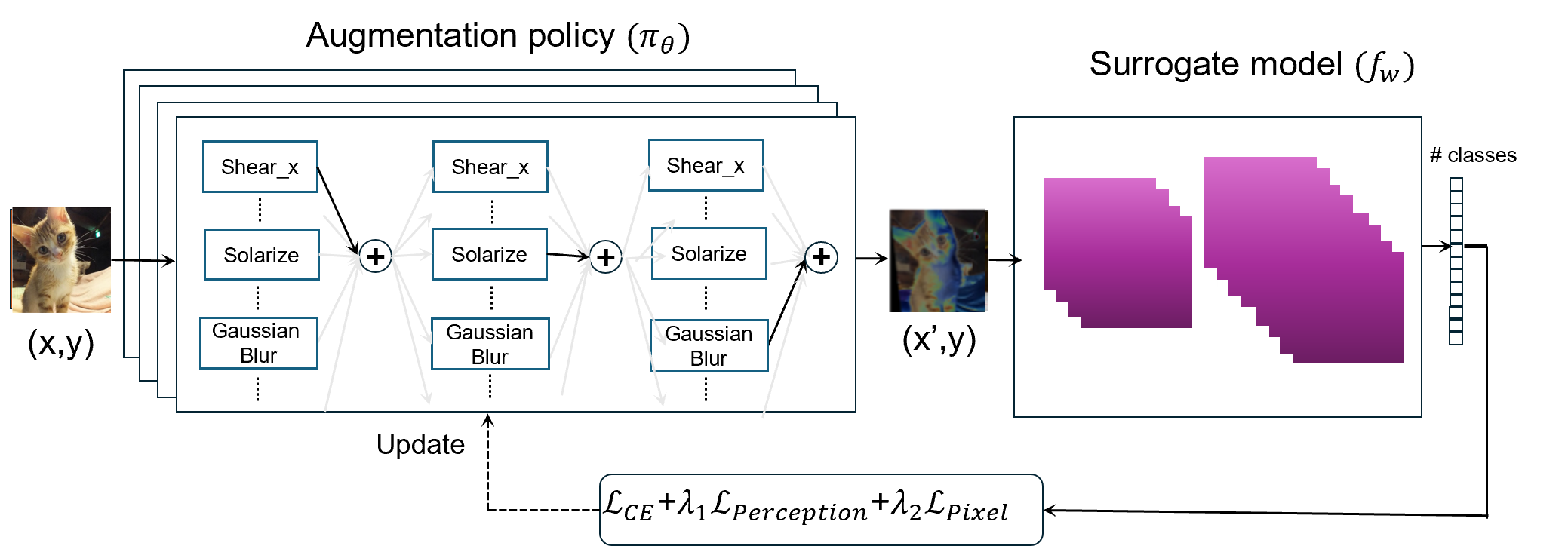}
    \caption{Overview of the gradient based adversarial augmentation framework (\techname). 
    Unlike existing augmentation-based attacks~\cite{hapuarachchi2024nonlinearTransformationsAgainst, qin2023learning, liu2023ImageShortcutSqueezing}, our search space is substantially expanded, comprising up to 18 augmentation operations spanning geometric transformations, color adjustment, Gaussian blur, and grayscale conversion; we select the operative subset per protection via grid search (Sec.~\ref{sec:gridsearch}). Each circle denotes the aggregated image, the weighted average of the candidate space's augmented images at each stage. Bold arrows represent the most dominant operations, explaining why the output image $x'$ is sheared, solarized, and blurred.}
    \label{fig:adv_aug_framework}
\end{figure*}

As illustrated in Fig.~\ref{fig:adv_aug_framework}, our framework adapts the identify-then-disrupt structure onto two components: (1) a surrogate model $f_w$, instantiated as ResNet-18, which performs identification by absorbing the protection's shortcut during fitting; and (2) an adversarial augmentation policy $\pi_\theta$, parameterized by $\theta$, which performs disruption by applying a learned transformation $T_\theta(x)$ to each unlearnable image before it is sent to $f_w$. Unlike UEraser~\cite{qin2023learning}, which entangles identification and disruption in a single training loop, {\techname} first fits the surrogate model once on a protected dataset to convergence; it then holds the surrogate fixed while learning the policy for breaking the underlying unlearnable pattern. This decoupling, together with the policy's differentiability, is what allows {\techname} to search a broad operation space end-to-end rather than the small, curated pools required by ISS~\cite{liu2023ImageShortcutSqueezing} and UEraser~\cite{qin2023learning}, or the sequential, combinatorial search used by nonlinear transformation~\cite{hapuarachchi2024nonlinearTransformationsAgainst}.

At each stage $s \in \{1,\dots,S\}$ of a sub-policy, every candidate operation is applied to the input, and the results are combined via a softmax-weighted sum over the stage's learnable weights $w_s$: $\sigma_\eta(w_s) = \mathrm{softmax}(w_s/\eta)$, where $\eta$ is a temperature parameter. During the policy training, this weighted combination is used directly, so the stage output is a differentiable function of $w_s$. Because softmax is differentiable, gradients flow from the surrogate model's loss back through every candidate operation, enabling the policy to be learned end-to-end via backpropagation~\cite{hataya2020faster}. At the neural network's training phase, a single operation is instead sampled from the same categorical distribution $\sigma_\eta(w_s)$, collapsing the weighted combination to a discrete transformation.
\subsubsection{Search space}\label{sec:search_space}
As summarized in Table~\ref{tab:summary_data_aug}, existing approaches either infer a fixed, curated operation set based on an image analysis of perturbation patterns~\cite{liu2023ImageShortcutSqueezing}, or restrict their search to a handful of hand-picked operations~\cite{qin2023learning}~\cite{hapuarachchi2024nonlinearTransformationsAgainst}. In contrast, we adopt the full sixteen-operation search space defined by AutoAugment~\cite{cubuk2018autoaugment}, and further add Gaussian blur and grayscale conversion. ISS~\cite{liu2023ImageShortcutSqueezing} shows that these two operations are effective against data protection perturbations. From this expanded search space, we determine the most effective subset for each protected dataset via a grid search over four candidate configurations rather than manual selection. Sec.~\ref{sec:gridsearch} details the grid search procedure, which we run independently for EMIN and NTGA, with the remaining four protections reusing EMIN's selected space.

\subsubsection{Gradient based Policy Search Process}\label{sec:gradient_search}
The search proceeds in two steps. First, the surrogate model $f_w$ is trained to convergence on the unlearnable dataset $D_u$, so that it absorbs the same shortcut signal a victim model would. A surrogate model that is briefly trained on a protected dataset or randomly initialized would provide a weak training signal for the policy to learn against, since the policy is trained to maximize the surrogate model's loss. Second, we train the policy $\pi_\theta$ such that its induced transformation $T_\theta$ maximizes the surrogate model's classification loss, while remaining close to the original image in both feature space and pixel space:
\begin{equation}
\theta^* \leftarrow \arg\min_{\theta} \;
- L_{\mathrm{ce}}\!\left(f_w\!\left(T_{\theta}(x)\right), y\right)
+ \lambda_1 R_{\mathrm{perc}}(\theta)
+ \lambda_2 R_{\mathrm{pix}}(\theta)
\label{eq:policy_objective},
\end{equation} where $R_{\mathrm{perc}}(\theta) = \left\lVert \phi\!\left(T_{\theta}(x)\right) - \phi(x) \right\rVert_2^2$ is a perceptual loss measured in the feature space of a pre-trained VGG-16 network, penalizing large feature-space distance between the augmented and original image so the policy does not destroy the underlying object content. $\phi(\cdot)$ denotes the 256-channel feature map extracted from the last ReLU activation of the third convolutional block, prior to max-pooling, of the VGG-16 network. The network is pretrained on ImageNet and kept frozen throughout policy search. $R_{\mathrm{pix}}(\theta) = \left\lVert T_{\theta}(x) - x \right\rVert_2^2$ is a pixel-space $\ell_2$ loss, penalizing large low-level deviations between the augmented and original image and complementing the perceptual term's high-level feature constraint. We directly optimize the objective function in Eq.~\ref{eq:policy_objective} by leveraging the differentiable DA pipeline and its gradient estimation of non-differentiable DA techniques~\cite{hataya2020faster}. This gradient based formulation addresses a scalability limitation inherent to UEraser's search strategy. UEraser-Max approximates its inner maximization by exhaustively exploring combinations of the fixed curated operation set~\cite{qin2023learning}, compensating for the imprecision of this combinatorial approximation by oversampling multiple augmented candidates per image and retaining only the highest-loss one. This 
strategy is tractable only because the search space is small; extending it up to the eighteen-operation space we adopt (Sec.~\ref{sec:search_space}) would require exploring combinatorially many more operation pairings, substantially increasing UEraser's search cost. By contrast, our policy $\pi_\theta$ optimizes Eq.~\ref{eq:policy_objective} directly using stochastic gradient ascent via the differentiable DA pipeline. This allows us to search the full eighteen-operation space without restricting it in advance or requiring manual selection based on a specific protection method's perturbation pattern.
\subsubsection{Search Space Selection via Grid Search}\label{sec:gridsearch}
The four candidate spaces ($S_1$–$S_4$, ranging from 16 to 18 operations) are nested subsets ($S_1 \subset S_2, S_3 \subset S_4$). We select an operative $(S,K)$ pair per protection empirically by jointly searching over the four spaces and $K\in\{2,3\}$ operations per sub-policy. We run $T=4$ independent trials for each $(S_j,K)$ combination and select $(S^*,K^*) = \arg\max_{j,k} \; \frac{1}{T}\sum_{t=1}^{T}
\mathcal{L}_{\text{ce}}\big(f_{w^*}(T_{\theta_{j,k,t}^*}(x)), y\big)$. This yields $(S^*,K^*)=(S_2,3)$ (base+blur) for EMIN and $(S_4,3)$ (base+blur+grayscale) for NTGA. The EMIN result supports the straight-through-estimator bias hypothesized in Sec.~\ref{sec:Discussion_limit}. AR, TAP, SEP, and DC reuse EMIN's selected space without an independent search, a decision we revisit in Sec.~\ref{sec:Discussion_limit} given AR and DC's underperformance relative to RS.

\subsection{Training with the Learned Augmentation Policy}\label{sec:max-loss_training}
We observed that standard training with the learned augmentation policy on $D_u$ obtains high training accuracy while clean test accuracy remains near 25\%. Inspection of the transformations applied during these epochs revealed that the policy was drawing sub-policies incapable of sufficiently removing the shortcut described in Sec.~\ref{sec:gradient_search}. Instead of learning generalizable features, the model continues fitting to these harmful shortcuts. 

We therefore adopt the two-phase training schedule of UEraser~\cite{qin2023learning}, using our learned sub-policies $\{S^{(1)}, \dots, S^{(L)}\}$
in place of the UEraser's fixed augmentation set. For the first $W$ epochs (error-maximizing phase), each training example $x_i$ is augmented $K$ times by randomly sampling $K$ sub-policies with replacement from $\{S^{(1)}, \dots, S^{(L)}\}$, producing candidates \{$T_\theta(x_i)^{(1)}, \dots, T_\theta(x_i)^{(K)}\}$; within each sub-policy, the operation at every stage is a sample from a categorical distribution over the learned softmax weights $Cat(\sigma_{\eta}(w_k))$. With a low temperature $\eta$, the distribution becomes non-uniform. Operations with higher probabilities are more likely to be chosen, but the sampling process remains random. The candidate incurring maximum per-sample loss under the current target model $f_v$, 
\begin{equation}
x_i^{*}
=
\arg\max_{k \in \{1,\ldots,K\}}
L\!\left(f\!\left(T_{\theta}(x_i)^{(k)}\right), y_i\right),
\label{eq:hardest_augmentation}
\end{equation}
is selected, and $f_v$ is updated by gradient descent on the batch-averaged max-loss:
\begin{equation}
v \leftarrow v - \eta \nabla_{v}
\left(
\frac{1}{B}
\sum_{i=1}^{B}
L_{\mathrm{ce}}\!\left(f_{v}(x_i^{*}), y_i\right)
\right)
\label{eq:model_update}
\end{equation}
For all subsequent epochs ($n > W$), $K$ is reduced to 1. Each image is augmented once via a single randomly sampled sub-policy, and training proceeds as minimizing the cross-entropy loss using the batch of those augmented images. This mirrors UEraser's own finding that error-maximizing augmentation is most valuable early in training, when the model must be forced past the unlearnable shortcuts, and can be relaxed to a cheaper single-sample pass once the shortcut has been sufficiently disrupted.
\subsection{Controlled Ablation}\label{sec:ablation}
To isolate the effect of gradient based search from the augmentation policy space itself, we compare two augmentation search strategies that differ only in how the maximization problem in Eq.~\ref{eq:policy_objective} is solved, holding the policy space, surrogate model architecture, and search budget fixed. In Sec.~\ref{sec:method}, \textbf{\techname} solves the maximization problem via the differentiable gradient ascent procedure. \textbf{RS} (Random Sampling) retains UEraser's original random $K$-sampling strategy, applied over the same expanded policy space, serving as the ablation control. Any performance gap between {\techname} and RS is therefore attributable to the difference between gradient guidance and random sampling as search strategies, rather than to the augmentation space or surrogate model architecture.

\section{Experimental Evaluation and Results}\label{sec:evaluation}

\subsection{Dataset and Models}
We evaluated {\techname} on CIFAR-10~\cite{krizhevsky2009learning}. We considered six unlearnable CIFAR-10 datasets generated by EMIN~\cite{EM}, AR~\cite{AR}, TAP~\cite{fowl2021adversarial}, NTGA~\cite{yuan2021neural}, SEP~\cite{DBLP:conf/iclr/ChenYC0QWH23}, and DC~\cite{deepconfuse}. All unlearnable perturbations are generated at a poison rate of 100\%. The perturbation magnitude is $8/255$. We used ResNet-18~\cite{he2016deep} as the surrogate model during the gradient based policy search process (Sec.~\ref{sec:gradient_search}). 

To assess the classification performance recovered by {\techname}, ResNet-18, ResNet-50~\cite{he2016deep}, and MobileNetV2~\cite{sandler2018mobilenetv2} were fitted on each protected dataset. We evaluated {\techname}'s effectiveness using the test accuracy of the trained models on clean test examples; higher clean test accuracy indicates greater effectiveness. We trained all target models for 200 epochs with a batch size of 128.

Within this setup, we compared three augmentation strategies: (1)~\textit{UEraser}~\cite{qin2023learning}; (2)~\textit{RS (Random Sampling)}, a control that samples augmentations randomly from {\techname}'s differentiable search space rather than performing gradient ascent, isolating the contribution of gradient guidance from the search space itself; and (3)~\textit{\techname}, our proposed gradient based adversarial augmentation framework. This evaluation method supports three following evaluation axes:
\begin{itemize}
    \item \textbf{Gradient guidance vs.\ Random sampling} (Sec.~\ref{sec:eval_gradient_vs_random}): comparing {\techname} against RS control under an identical search space and surrogate model, to isolate the benefit of the gradient guided search over random sampling.
    \item \textbf{Cross data protection transfer of the learned policy} (Sec.~\ref{sec:eval_cross_data_protection_transfer}): since the policy is searched using only the EMIN based surrogate, evaluation on CIFAR-10 protected by EMIN constitutes a matched comparison, while evaluation on the remaining protected datasets constitutes a transfer setting.
    \item \textbf{Cross model transfer of the learned policy} (Sec.~\ref{sec:eval_cross_model_transfer}): since the policy is searched using a ResNet-18 surrogate, evaluation on ResNet-18 target models constitutes matched-architecture performance, while ResNet-50 and MobileNetV2 constitute cross-architecture transfer.
\end{itemize}
\subsection{Training settings}
\textbf{DA policy search settings:} The policy was searched on a 5000 image subset of the respective protection CIFAR-10 training set, augmented with standard RandomCrop and RandomHorizontalFlip. Following the search space selection procedure in Sec.~\ref{sec:gridsearch}, we used $(S^*,K^*)=(\text{base+blur}, 3)$ for EMIN, AR, TAP, SEP, and DC, and $(S^*,K^*)=(\text{base+blur+grayscale}, 3)$ for NTGA. The policy is trained for 10 epochs with batch size 128, using Adam ($\beta_1=0$, $\beta_2=0.999$, learning rate 0.001), with $\lambda_{1}=\lambda_{2}=0.1$ in Eq.~\ref{eq:policy_objective}.

\textbf{DNN model training settings:} For the optimizer, we used SGD with a learning rate of 0.01, a momentum of 0.9, and a weight decay of $5\times10^{-4}$. We varied the learning rate via a cosine annealing schedule over the full 200 epoch run, and we clipped the gradients to a maximum norm of 5.0. Following UEraser's error-maximizing warmup schedule, we applied the $K$-sample maximization procedure across all three adversarial augmentation search strategies for the first $W=50$ epochs, after which training switched to single-pass augmentation for the remaining 150 epochs.
\subsection{Results}
Table~\ref{tab:grad_search} reports clean test accuracies (\%) on CIFAR-10 with ResNet-18, comparing our gradient based adversarial augmentation ({\techname}) against a random sampling adversarial augmentation (RS) using the six unlearnable datasets. We also evaluated UEraser~\cite{qin2023learning}, using its fixed augmentation policies as implemented in APBench~\cite{qin2024apbench}; unlike {\techname}, UEraser's policy is fixed by design and involves no search phase, so it incurs no equivalent search time cost. All three DA based attack methods were trained under the identical two phase training procedure described in Sec.~\ref{sec:max-loss_training}. We held the training pipeline constant so that the only difference between the three methods is the adversarial augmentation-search mechanism, including gradient ({\techname}), random (RS), or heuristic (UEraser). Note that this yields different absolute accuracies than UEraser's original paper, which uses its own training configuration. Our setup instead prioritizes a controlled comparison over reproducing UEraser's published numbers exactly.
\begin{table}[htbp]
\caption{ Clean test accuracies(\%) of {\techname} and a random-sampling adversarial
augmentation (RS) on CIFAR-10.}
\begin{center}
{\setlength{\tabcolsep}{2pt}
\begin{tabular}{p{1.20cm}|c|c|c|c|c|c|c|c}
\toprule
    Attacks & Clean & EMIN  & AR & NTGA & SEP & TAP & DC \\ \hline
{\techname}
     & 95.57 & 80.0 & 22.10 & 54.95 & 78.06 & 80.0  & 70.34 \\
RS
     & 94.78 & 77.29 & 48.01 & 80.28 & 70.76 & 77.29 & 90.44\\
{UEraser}
 & 91.93 & 83.88 & 89.36 & 90.21 & 89.74  & 88.39  & 89.66 \\
\bottomrule
\end{tabular}}
\label{tab:grad_search}
\end{center}
\raggedright
\footnotesize
$^{\dagger}${\techname} results use the EMIN-selected search space and
sub-policy length, $(S^*,K^*)=$(base+blur, 3), for all columns
(EMIN, AR, NTGA, SEP, TAP, DC) (Sec.~\ref{sec:eval_cross_data_protection_transfer}).
 \end{table}
\subsubsection{Gradient guidance vs. Random sampling}\label{sec:eval_gradient_vs_random}
The surrogate model $f_w$ was trained once on the EMIN dataset. The training pipeline and evaluation conditions are identical for both {\techname} and RS, making this a fair, controlled ablation comparison. Under this matched setting, {\techname} achieves 80.0\% versus RS's 77.29\%. Thus, {\techname} finds a more effective adversarial augmentation policy than RS when we held  the policy space, search budget, and training pipeline fixed. UEraser's fixed policy reaches 83.88\% under the same training pipeline, outperforming both counterparts on EMIN, which we will discuss in Sec.~\ref{sec:Discussion}.
To test whether {\techname}'s advantage over RS generalizes beyond EMIN, we additionally searched a policy on the NTGA dataset, using the same procedure as for EMIN (Sec.~\ref{sec:gradient_search}), and evaluated the NTGA matched policy against RS under the same surrogate and evaluation conditions. Table~\ref{tab:ntga_matched} demonstrates that {\techname} outperforms RS by 5.6–16.8\% across ResNet-18, ResNet-50, and MobileNetV2, mirroring the direction of the EMIN result. This suggests {\techname}'s advantage over the random sampling is not specific to EMIN, but requires the policy to be searched against the data protection approach it is evaluated on.
\begin{table}[htbp]
\caption{Clean and NTGA-protected test accuracies (\%) under an NTGA matched policy search, i.e., where the surrogate and policy are trained specifically on NTGA dataset rather than transferred from the EMIN trained surrogate.}
\begin{center}
\begin{tabular}{l|c|c|c}
\toprule
Model & {\techname} & RS & UEraser \\
\midrule
ResNet-18 & 85.88 & 80.28 & 90.21 \\
ResNet-50 & 86.68 & 78.27 & 90.75 \\
MobileNet-V2 & 83.28 & 66.45 & 88.83 \\
\bottomrule
\end{tabular}
\label{tab:ntga_matched}
\end{center}
\raggedright
\footnotesize
$^{\dagger}$Policy searched specifically on NTGA dataset using NTGA's own grid-search-selected space and sub-policy length, $(S^*,K^*)=$(base+blur+grayscale, 3) (Sec.~\ref{sec:gridsearch}),
distinct from the EMIN-selected $(S^*,K^*)$ used in Table~\ref{tab:grad_search}.
\end{table}

\subsubsection{Cross data protection transfer of the learned policy}\label{sec:eval_cross_data_protection_transfer}
For the remaining data protection approaches (AR~\cite{AR}, NTGA~\cite{yuan2021neural}, SEP~\cite{DBLP:conf/iclr/ChenYC0QWH23}, TAP~\cite{fowl2021adversarial}, DC~\cite{deepconfuse}), the policy searched against the surrogate model trained on the EMIN dataset was applied to a different protected dataset, which makes it a cross data protection transfer setting rather than a matched ablation. Under this setting, the pattern reverses: RS substantially outperforms {\techname} on AR (48.01\% vs. 22.10\%), NTGA (80.28\% vs. 54.95\%), and DC (90.44\% vs. 70.34\%), while {\techname} retains an advantage on SEP (78.06\% vs. 70.76\%). A matched policy searched specifically on NTGA closes and reverses this gap (Sec.~\ref{sec:eval_gradient_vs_random}, Table~\ref{tab:ntga_matched}), suggesting the transfer failure reported here reflects policy protected dataset mismatch rather than a fundamental limitation of {\techname}. On the other hand, UEraser's fixed policy, which was never specialized to any perturbation pattern, outperforms both {\techname} and RS on every one of these five attacks (89.36\%–90.21\%), consistent with the interpretation that specialization to EMIN's shortcut structure is actively counterproductive when the evaluation datasets diverge from it. This indicates the learned policy's benefit depends on the similarity between the surrogate's training dataset and the evaluation dataset, rather than a general property of the gradient based search. We discuss the implications of this transfer gap and UEraser's consistent advantage across the non-EMIN datasets in Sec.~\ref{sec:Discussion}.
\subsubsection{Cross model transfer of the learned policy}\label{sec:eval_cross_model_transfer}
Table~\ref{tab:arch_transfer} reports clean test accuracies (\%) for ResNet-50 and MobileNet-V2 models, trained on unlearnable CIFAR-10 across the same six attacks (EMIN~\cite{EM}, TAP~\cite{fowl2021adversarial}, NTGA~\cite{yuan2021neural}, SEP~\cite{DBLP:conf/iclr/ChenYC0QWH23}, AR~\cite{AR}, DC~\cite{deepconfuse}), following the same three-way comparison ({\techname}, RS, and UEraser) and the same training pipeline (Sec.~\ref{sec:max-loss_training}). The adversarial augmentation policy for {\techname} was searched once using a ResNet-18 surrogate (Sec.~\ref{sec:gradient_search}) and then applied unmodified when training each target architecture. This tests whether a policy learned against one architecture's shortcut representation can transfer across target models.
\begin{table}[htbp]
\caption{Clean test accuracies (\%) of different architectures (ResNet-50 and MobileNet-V2) on CIFAR-10.}
\begin{center}
\begin{tabular}{c|c|cc}
\toprule
     Attacks &
    & ResNet-50
    & MobileNet-V2
    \\
    \midrule
    \multirow{7}{*}{\techname}
    & Clean
    & 96.15  & 91.59 \\
    & EMIN
    &  83.09 & 76.91 \\
    & TAP
    & 83.17  & 87.99 \\
    & NTGA
    &  57.64 & 48.02 \\
    & SEP
    & 78.42  & 72.16\\
    & AR
    & 20.91 & 17.59\\
    & DC
    & 78.48  & 78.87 \\
    \midrule
    \multirow{7}{*}{RS}
    & Clean
    & 96.07 & 94.88 \\
    & EMIN
    & 83.21 & 76.78  \\
    & TAP
    & 83.54 & 76.78  \\
    & NTGA
    & 78.27 & 66.45  \\
    & SEP
    & 77.80 & 70.62  \\
    & AR
    & 46.15 & 48.05  \\
    & DC
    & 89.94 & 80.77  \\
    \midrule
    \multirow{7}{*}{UEraser}
    & Clean
    & 92.44 & 90.22 \\
    & EMIN
    & 86.13 & 81.31  \\
    & TAP
    & 89.10 & 87.5 \\
    & NTGA
    & 90.75 & 88.83  \\
    & SEP
    & 90.59 & 87.55  \\
    & AR
    & 89.87 & 87.65  \\
    & DC
    & 90.99 & 88.33  \\
\bottomrule
\end{tabular}
\label{tab:arch_transfer}
\end{center}
\end{table}
When we trained the surrogate model on the EMIN dataset, {\techname} and RS performed equally on both architectures. While {\techname} achieves 83.09\% and 76.91\% on ResNet-50 and MobileNet-V2, respectively, RS obtains 83.21\% on ResNet-50 and 76.78\% on MobileNet-V2. Neither shows the clear advantage seen on ResNet-18 (Sec.~\ref{sec:eval_gradient_vs_random}); instead, UEraser achieves the highest accuracy on both architectures, with 86.13\% on ResNet-50 and 81.31\% on MobileNet-V2.
On the cross model transfer scenario, the pattern from the cross data protection transfer setting (Sec.~\ref{sec:eval_cross_data_protection_transfer}) repeats. RS outperforms {\techname} on NTGA (ResNet-50: 78.27\% vs. 57.64\%; MobileNet-V2: 66.45\% vs. 48.02\%) and AR (ResNet-50: 46.15\% vs. 20.91\%; MobileNet-V2: 48.05\% vs. 17.59\%), while results are closer on SEP and TAP~\cite{fowl2021adversarial}. UEraser again outperforms both {\techname} and RS across both architectures.

Taken together, Tables~\ref{tab:grad_search} and~\ref{tab:arch_transfer} show the effectiveness gap across the three data augmentation attack (Sec.~\ref{sec:eval_cross_data_protection_transfer}) is not specific to ResNet-18. It persists across target architectures, suggesting the cause lies in the specialization of the searched policy to EMIN's noise pattern rather than in any interaction with a particular architecture.
\section{Discussion}\label{sec:Discussion}
\subsection{Adaptive data protection}
{\techname} relies on the gradient of the surrogate model's loss function to update the policy weight for each operation at a given stage. Specifically, the gradient is designed to significantly increase the loss function. A defender aware of {\techname}'s search space could, in principle, harden a protection method's shortcut against specific operations. For instance, a data owner can construct a perturbation that remains effective under grayscale conversion, the operation Liu et al.~\cite{liu2023ImageShortcutSqueezing} identify as broadly effective against protection methods and {\techname} includes in its search space (Sec.~\ref{sec:search_space}). We argue {\techname} is more robust to such adaptive protection method than a fixed-policy defense. Gradient ascent on the policy weights follows
\begin{equation}
\frac{\partial L}{\partial w_k} = \frac{\partial L}{\partial T_{\theta}(x)} \cdot \frac{\partial T_{\theta}(x)}{\partial w_k},
\end{equation} where the first term reflects the surrogate model's loss sensitivity to a given change in the augmented image. If the defender constructs a shortcut that is invariant to a specific operation, the surrogate will predict the shortcut correctly regardless of how you slide the image toward more extreme magnitude values; hence, the surrogate model's loss barely changes, and $\frac{\partial L}{\partial T_{\theta}(x)}$ for that operation goes to near zero. However, any operation the defender does not anticipate will still produce a large gradient and the policy's weight distribution concentrates over those operations whose gradient stays large, limiting the effectiveness of the adaptive defender. Therefore, it remains challenging for the adaptive defender to improve its robustness against the transformations {\techname} applies. 

\subsection{Limitations}\label{sec:Discussion_limit}
\textbf{The straight-through gradient estimator bias:} {\techname} achieves 80.00\% clean test accuracy, substantially exceeding UEraser's poorly ordered policy combinations (69.24\%, e.g., PlasmaTransform followed by ChannelShuffle) but trailing their best performing ordering (83.88\%, ChannelShuffle, PlasmaTransform, and TrivialAugment) under the EMIN protected dataset with ResNet-18. We hypothesize this shortfall stems from accumulating bias in the straight-through estimator's gradient approximation through the non-differentiable augmentation operations, which may compound over successive gradient ascent steps and yield a less precise solution to the inner maximization than UEraser's combinatorial search over its fixed, curated operation set.

\textbf{Cross data protection transfer issue:} 
We interpret the policy $\pi_\theta$ as a generator producing adversarial augmentations tuned to disrupt the specific shortcut structure the EMIN-trained surrogate $f_w$ has learned. This framing is consistent with the strong cross model transfer observed in Sec.~\ref{sec:eval_cross_model_transfer}, where the shortcut is held fixed and only the target architecture varies. It does not, however, predict success under cross data protection transfer case (Sec.~\ref{sec:eval_cross_data_protection_transfer}), where the underlying shortcut structure itself changes across protection methods. This structural account is further supported by prior findings on the linear separability of data protection perturbations. EMIN~\cite{EM} and TAP~\cite{fowl2021adversarial} perturbations exhibit strong linear separability while NTGA's noise, though still separable, is comparatively less so. AR's perturbation, by contrast, is not linearly separable~\cite{sandoval2023can}. This ordering corresponds closely to {\techname}'s cross data protection transfer performance. TAP, whose shortcut structure most closely resembles EMIN's, achieves clean accuracy (80.0\%). NTGA, with weaker shared structure, transfers only partially (54.95\%, Table~\ref{tab:grad_search}). AR, sharing no comparable linear structure with EMIN, exhibits the largest transfer failure (22.10\%). This pattern indicates that the effectiveness of GAA’s policy transfer across protection methods depends on how structurally similar their underlying shortcuts are.  We do not have linearity separability status for SEP~\cite{DBLP:conf/iclr/ChenYC0QWH23} and DC~\cite{deepconfuse}, so we cannot extend this account to those cases. SEP's comparatively strong transfer (78.06\%) and DC's more moderate transfer (70.34\%) remain unexplained by this hypothesis.

This account is further supported by a direct test. The {\techname} policy searched specifically on the NTGA dataset (Table~\ref{tab:ntga_matched}) outperforms RS by 5.6\%, 8.41\%, and 16.8\% across ResNet-18, ResNet-50, and MobileNetV2, respectively. These results confirm that {\techname}'s advantage over random sampling persists once the policy protected dataset mismatch is removed. Thus, the policy protected dataset mismatch causes the effectiveness gap rather than {\techname}'s weakness. Verification for the remaining attacks, including AR, SEP, TAP, and DC, is left to future work.

Critically, neither limitation affects the paper's core finding. The gradient based search ({\techname}) vs. the random sampling ablation (Sec.~\ref{sec:ablation}), conducted under identical policy space and search budget, shows that {\techname} still outperforms random sampling when all other factors are held fixed.

\section{Conclusion}\label{sec:Conclusion}
We presented {\techname}, a gradient based adversarial augmentation framework that learns to mitigate unlearnable data methods by searching a space of up to 18 operations, from which a protection specific operative subset is selected via grid search, rather than the heuristic search strategies introduced by nonlinear transformation~\cite{hapuarachchi2024nonlinearTransformationsAgainst}, ISS~\cite{liu2023ImageShortcutSqueezing}, and UEraser~\cite{qin2023learning}. Under a controlled ablation that holds the policy space, surrogate model, and training pipeline fixed, {\techname}'s gradient based search consistently outperforms random sampling over the same search space (80.0\% vs. 77.29\% clean test accuracy on EMIN protected CIFAR-10 for ResNet-18 and 83.28\% vs. 66.45\% on NTGA protected CIFAR-10 for MobileNet-V2), demonstrating that gradient guidance, not the search space expansion, is responsible for the improvement. This advantage does not extend to the cross data protection transfer setting. Applying {\techname}'s learned policy, which specializes to EMIN's shortcut structure to AR, NTGA, SEP, TAP, and DC shows that its benefit over random sampling diminishes. We argue that this effect corresponds to the degree of linear separability in perturbations. By contrast, the learned policy's cross model transferability remains stable across ResNet-18, ResNet-50, and MobileNetV2. Future work includes constructing an adaptive defender that explicitly targets {\techname}'s search space to empirically test the robustness 
and evaluating {\techname} under different perturbation magnitudes and on additional datasets such as SVHN, CIFAR-100, and ImageNet-1k.

\bibliographystyle{IEEEtran}
\bibliography{references.bib}


\end{document}